# End-to-End Learning Via A Convolutional Neural Network For Cancer Cell Line Classification

**Darlington A. Akogo**

minoHealth AI Labs

**Xavier-Lewis Palmer**

Old Dominion University, minoHealth AI Labs

**July 2018**

**Abstract**

Computer Vision for automated analysis of cells and tissues usually include extracting features from images before analyzing such features via various Machine Learning and Machine Vision algorithms. We developed a Convolutional Neural Network model that classifies MDA-MB-468 and MCF7 breast cancer cells via brightfield microscopy images without the need of any prior feature extraction. Our 6-layer Convolutional Neural Network is directly trained, validated and tested on 1,241 images of MDA-MB-468 and MCF7 breast cancer cell line in an end-to-end fashion, allowing a system to distinguish between different cancer cell types. The model takes in as input imaged breast cancer cell line and then outputs the cell line type (MDA-MB-468 or MCF7) as predicted probabilities between the two classes. Our model scored a 99% Accuracy.

## Introduction

Convolutional Neural Networks were developed initially in the 1980s and were called Neocognitron (Fukushima, 1980, 1983, 1987). They are broadly part of a wide set of models called *Multi-Stage Hubel-Wiesel Architectures.* In 1989, LeNet-5 was introduced which simplified the architecture and used the Backpropagation algorithm to train the entire architecture in a supervised fashion (Yann LeCun, 1989). The architecture was successful for tasks such as Optical Character Recognition and Handwriting Recognition. Convolutional Neural Networks have been an important aspect of Deep Learning in recent years. They were mainly responsible for the re-emergence and popularity of Neural Networks. The work of Alex Krizhevsky and Ilya Sutskever which won the ImageNet Large Scale Visual Recognition Competition in 2012 (ILSVRC-2012) was disruptive in the Artificial Intelligence, Machine Learning and Computer Vision community (Alex Krizhevsky et al., 2012). Since then Convolutional Neural Networks have been heavily applied to all sorts of problems, from various Object Detection and Image Segmentation problems (Liang-Chieh Chen et al., 2014, J. Redmon et al., 2015, S. Ren et al., 2015 ) and to specific domains like Medical Image Analysis (Shadi Albarqouni et al., 2016, Mark J. J. P. van Grinsven et al., 2016, Lin Yang et al., 2017, Andre Esteva et al., 2017).

Their effectiveness can be attributed to their ability to handle translation invariances in images by relying on shared weights and exploit spatial locality by enforcing a local connectivity pattern between neurons of adjacent layers. We chose them for this reason, knowing we wanted a model that could visually detect and differentiate between different breast cancer cell lines such as MDA-MB-468 and MCF7 in an end-to-end fashion. According to the American Cancer Society, breast cancer is the leading diagnosed cancer for American women, not including skin cancer, with more than 250,000 new cases of invasive cases expected and more than 40,000 deaths as of 2017, making it an important target to address stateside (DeSantis et al 2017). This is especially the case in developing regions abroad, where care is less accessible, especially due to a lack of sophisticated equipment, reagents, and more may hinder detection. In efforts to diagnose and treat cancer, tools that can assist less equipped labs are increasingly important. Within, we present a tool that can distinguish between images of cell lines via brightfield microscopy, without additional preparation, that may assist automated detection tools and diagnoses.

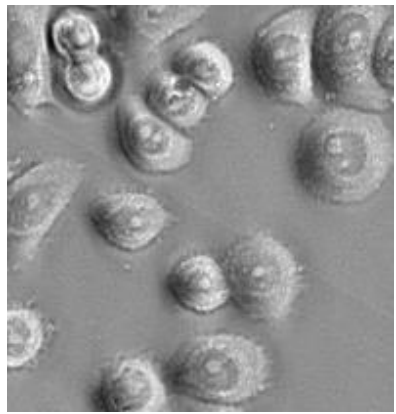 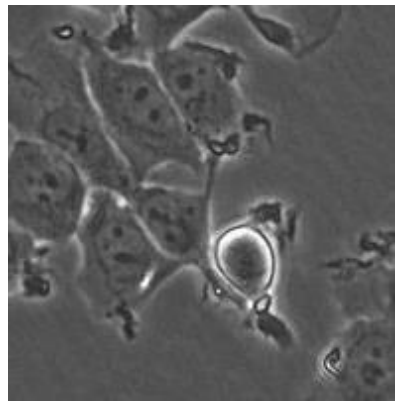

MDA-MB-468 sample            MCF7 sample

**Figure 1**. Samples of MDA-MB-468 and MCF7 breast cancer cell that were used in training the Convolutional Neural Network.

## Data

We used a collection of 1,241 grayscale images of MDA-MB-468 and MCF7 breast cancer cell for training, validation and testing our model. Sample images from the dataset can be seen in Figure 1 above. Our dataset contains 664 MDA-MB-468 breast cancer cell images and 577 MCF7 breast cancer cell as shown in Figure 2 below. MDA-MB-468 cells and MCF7 cells were cultured and then placed into 3 separate 6-well cell plates and imaged at 400X via brightfield microscopy. Images were separated into brightly and dimly lit categories and then tiled into 128 x 128 pixel images for analysis. Given the broadness of the wells and cell positions imaged, lighting differed, adding a challenge to the process, which can reflect practical research realities and variations, leading to difficulty in automated systems properly detecting cells.

The dataset was split into *Training set*, *Validation set* and *Testing set* with a *8:1:1 ratio* (995, 123, 123), respectively. The dimensions of all the images were reshaped to 128 by 128 pixels. The images were then transformed by Standardization so our image pixel values that'd act as inputs to our model would have a similar range for more stable gradients during training. To increase the variation in our dataset to ensure our trained model generalizes beyond its training data. We further augmented the images with random horizontal flips, 5° rotations, width shifts, height shifts and zooms.

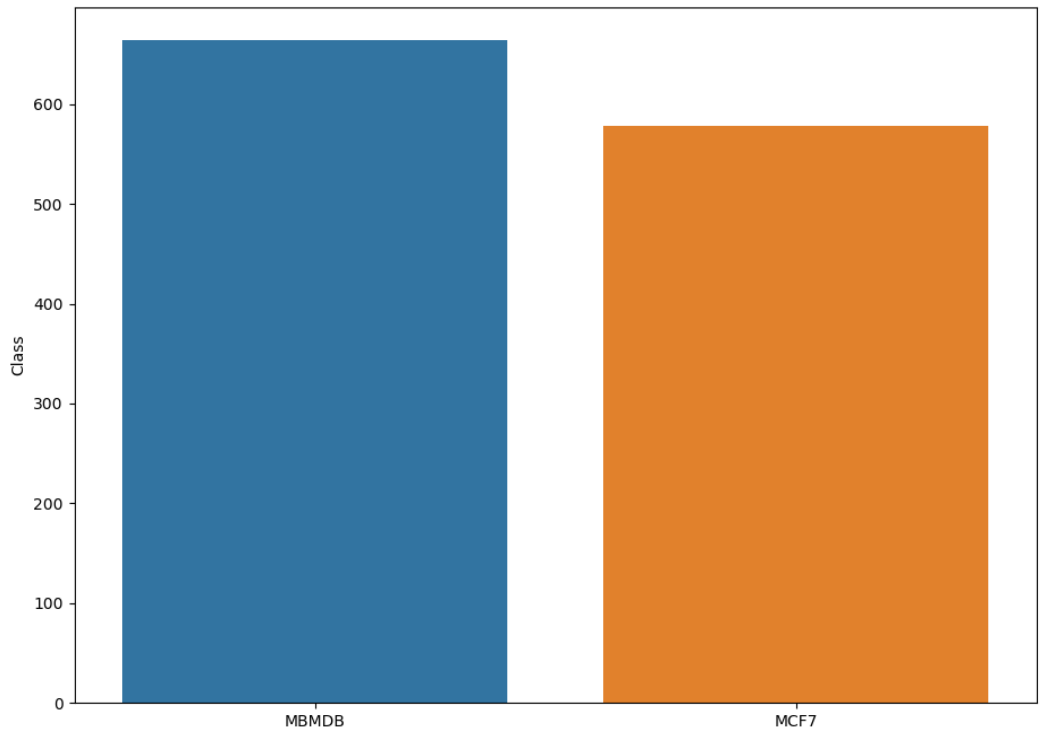

**Figure 3**. The complete dataset contains 664 MDA-MB-468 breast cancer cell images and 577 MCF7 breast cancer cells.

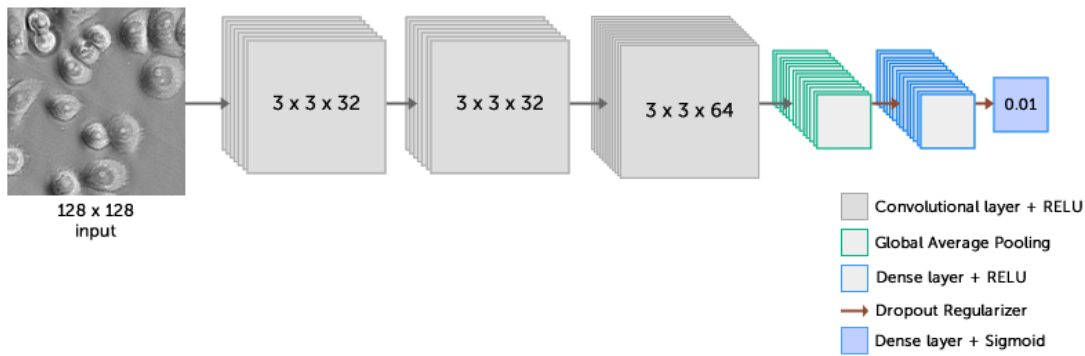

**Figure 4**. The architecture of our Convolutional Neural Network. Based on the ScaffoldNet architecture. The model propagates the input image through its 4 hidden layers and then outputs predicted probabilities between 0 and 1, represent both MDA-MB-468 and MCF7 classes respectively, with the threshold being 0.5. This example accurately outputs 0.01, which represents MDA-MB-468 cell class.

## Model Architecture

We use a 6-layer Convolutional Neural Network trained and tested on 1,241 grayscale images of MDA-MB-468 and MCF7 breast cancer cells. Our Convolutional Neural Network architecture is based on the **ScaffoldNet** architecture (Akogo DA and Palmer XL., 2018). As shown in Figure 4, our network ScaffoldNet starts with two 2-Dimensional Convolutional layers with a 3 x 3 kernel size and 32 output filters with the first as its input layer. Then followed by a single 2-Dimensional Convolutional layers also with a 3 x 3 kernel size and 64 output filters. We then introduce a 2-Dimensional Global Average Pooling layer to reduce the spatial dimensions of our tensor (Min Lin et al., 2013). Global Average Pooling performs Dimensionality Reduction to minimize overfitting by turning a tensor with dimensions $h \times w \times d$ into $1 \times 1 \times d$ which is achieved by reducing each $h \times w$ feature map to a single number simply by taking the average of all $hw$ values. To further prevent overfitting, we then add a Dropout Regularizer with a fraction rate of 0.5 (Srivastava et al., 2014). And then we introduce a 32 unit densely-connected Neural Network layer into our network architecture, followed by another Dropout Regularizer with a 0.5 fraction rate. Our final output layer is a single unit dense Neural Network layer.

All Convolutional and Densely-connected layers except the output layer use the Rectified Linear Unit (RELU) activation function:

$$f(x) = \max(0, x)$$

where $x$ is the input to a neuron (Hahnloser et al., 2000).

The final single neuron output layer uses a Sigmoid activation function:

$$S(x) = \frac{1}{1+e^{-x}}$$

where $x$ is the input to a neuron and $e$ is the natural logarithm base (also known as Euler's number)

Our Convolutional Neural Network is trained end-to-end with the first-order gradient-based optimization algorithm, Adam, using the standard parameters($β1$ = 0:9 and $β2$ = 0:999) (Kingma and Ba, 2014). And we use the Cross-Entropy loss function for Binary Classification:

$$-(y \log(p) + (1 - y) \log(1 - p))$$

where log - the natural log
$y$ - binary indicator (0 or 1) if class label c is the correct classification for observation o
$p$ - predicted probability observation

We train our model using mini-batches of 32. We use a learning rate($α$) of 0.001, and pick the model with the lowest validation loss.

**Model Training and Validation**

Using the Training set (995 images), our Convolutional Neural Network was trained with Adam optimization algorithm. *Cross-Entropy loss function* and the *Accuracy Classification Score* were used as metrics. The Accuracy score formula is;

$$\text{accuracy}(y, \hat{y}) = \frac{1}{n_{\text{samples}}} \sum_{i=0}^{n_{\text{samples}}-1} 1(\hat{y}_i = y_i)$$

where the $\hat{y}_i$ is the predicted output for $i$-th sample $y_i$ is the (correct) target output computed over $n_{\text{samples}}$

Our model was trained in 8 epochs and its hyperparameters tuned using the Validation set (123 images). After just the first epoch, our model had the following performance results on the Validation set;

Accuracy score: 94.31%

Cross-Entropy loss: 0.1801

After the 8th epoch, our model's final performance results on the Validation set were;

Accuracy score: 98.37%

Cross-Entropy loss: 0.0934

**Model Testing and Results**

   After all training and validation, we finally evaluated our model on the Test set (123 images). The Test set is the final evaluation for a model and changes are not made to the model after the results.

ScaffoldNet's final performance results on the Test set were;

Accuracy score: **99.00%**

Cross-Entropy loss: **0.0926**

   From the results of the final evaluation, we can tell that our model generalizes well and doesn't overfit, the high accuracy performance on the Validation set is consistent with the evaluation results on the Test set.

   To further evaluate our Convolutional Neural Network's output quality, we use the Receiver Operating Characteristic (ROC) metric and its Area Under Curve (AUC) score. We use the ROC curve to plot our model's true positive rate on the Y axis, and false positive rate on the X axis. Our Convolutional Neural Network classifier has a near perfect AUC score of 0.98 as shown in the plot below, in Figure 4.

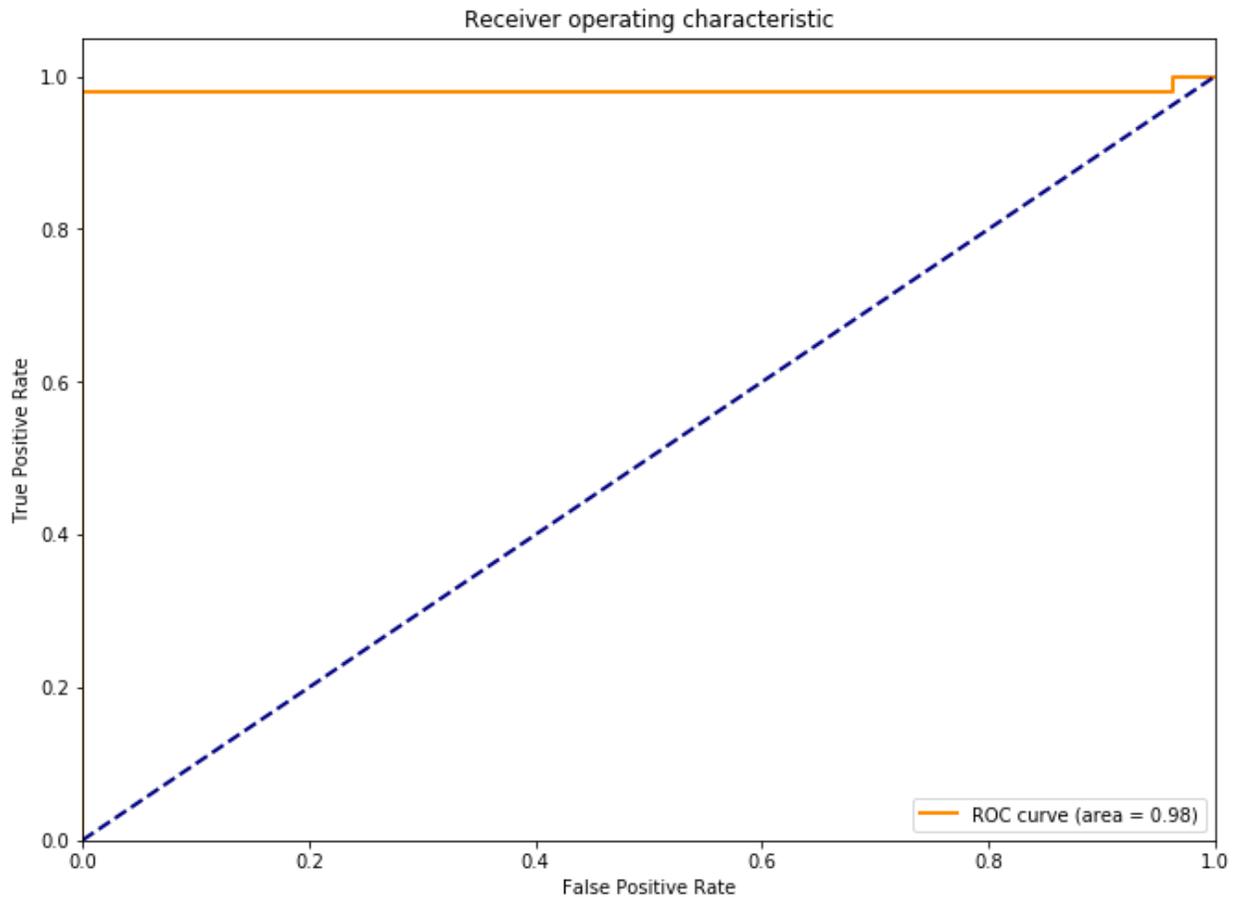

**Figure 4**. As seen on the curve, our Convolutional Neural Network Classifier has a high AUROC score of 0.98.

**Related Work**

Some earlier works exist within the domain of Computer Vision for automated analysis of cells and tissues. Some of such works segment various cells from each image then extract features like size and shape from such cells. These extracted features are then used to train Machine Learning models or further analyzed by other Machine Vision algorithms. This include examples where they extract features from segmented blood cells and then classify them via Multilayer Perceptrons (Wei Lin et al., 1998). Other examples include grading of cervical intraepithelial neoplasia by extracting geometrical features that are analyzed using a combination of computerized digital image processing and Delaunay triangulation analysis (Keenan SJ, et al., 2000). And localization of sub-cellular components via threshold adjacency statistics which are then analyzed by support vector machine (Hamilton NA et al., 2007). Others compare extracted features and raw pixel densities analyzed via Bayesian Classifier, K-Nearest Neighbors, Support vector machine, and Random Forest (Timothy BL et al., 2016). Unlike all these works, we used

Deep Learning in a End-to-End fashion, where we train our Convolutional Neural Network to directly analyze raw pixel values without any need for feature extraction. This drastically simplifies the process of developing automated Computer Vision systems for cell and tissue analysis. By eliminating feature extraction, Computer Vision system can then fully and truly learn important regularities pertaining cells themselves rather than being limited by rules via extracted features we create.

**Conclusion and Outlook**

We developed a Convolutional Neural Network that accurately classifies MDA-MB-468 and MCF7 breast cancer cells after being trained on 995 brightfield breast cancer cell images, validated with 123 brightfield breast cancer cell images, and then tested on 123 brightfield breast cancer cell images. The Convolutional Neural Network performed well, with a 99% Accuracy score and 0.98 AUC score, indicating reliability for classification purposes. We believe that this system holds promise for expansion into other cancerous and normal cell lines of other diseases cases as may be reflected in upcoming work. More importantly it can potentially help lower barriers for care in less equipped labs.